\documentclass[10pt,twocolumn,letterpaper]{article}

\usepackage{icb}
\usepackage{times}
\usepackage{epsfig}
\usepackage{graphicx}
\usepackage{amsmath}
\usepackage{amssymb}
\usepackage{float}

\icbfinalcopy 

\ificbfinal\fi
\begin{document}

\title{Fingerprint Distortion Rectification using Deep Convolutional Neural Networks}

\author{Ali Dabouei, Hadi Kazemi, Seyed Mehdi Iranmanesh, Jeremy Dawson, Nasser M. Nasrabadi\\
West Virginia University\\
{\tt\small \{ad0046, hakazemi, seiranmanesh\}@mix.wvu.edu, \{jeremy.dawson, nasser.nasrabadi\}@mail.wvu.edu}
}

\maketitle
\thispagestyle{empty}

\begin{abstract}
    Elastic distortion of fingerprints has a negative effect on the performance of fingerprint recognition systems. This negative effect brings inconvenience to users in authentication applications. However, in the negative recognition scenario where users may intentionally distort their fingerprints, this can be a serious problem since distortion will prevent recognition system from identifying malicious users. Current methods aimed at addressing this problem still have limitations. They are often not accurate because they estimate distortion parameters based on the ridge frequency map and orientation map of input samples, which are not reliable due to distortion. Secondly, they are not efficient and requiring significant computation time to rectify samples.  In this paper, we develop a rectification model based on a Deep Convolutional Neural Network (DCNN) to accurately estimate distortion parameters from the input image. Using a comprehensive database of synthetic distorted samples, the DCNN learns to accurately estimate distortion bases ten times faster than the dictionary search methods used in the previous approaches. Evaluating the proposed method on public databases of distorted samples shows that it can significantly improve the matching performance of distorted samples.  
\end{abstract}

\section{Introduction}

The fingerprint is one of the most important biometric modalities due to its uniqueness and easy acquisition process. 
Leveraged by rapid advances in sensor technologies and matching algorithm development, automatic fingerprint recognition has been widely adopted as a highly-accurate identification method. 
The operation of a typical fingerprint recognition system consists of three main steps. In the preprocessing step, a raw fingerprint is enhanced to reduce noise, connect broken ridges and separate joined ridges. In the second step, exact ridge patterns are processed to extract local features, namely minutiae, from the enhanced image. In the final step, a match score between two fingerprint features is calculated by analyzing properties of minutiae (location, orientation, etc.) using local and global relationships between them.
	
In past decades, algorithms for fingerprint matching have advanced rapidly, resulting in the development of numerous and varied commercial fingerprint recognition systems. These algorithms have very high performance in identifying clean samples \cite{cappelli2006}, but often fail in identifying samples which are distorted. Consequently, recognizing dirty fingerprints is a challenging problem for fingerprint recognition systems. Most of the fingerprint matching algorithms are based on calculating the relative properties between features within a fingerprint, and matching them with other fingerprints. However, distortion that can occur during the collection process changes the relative properties of fingerprint features and causes a notable decrease in recognition performance \cite{FVC2006}.

There are two main types of recognition scenarios. 
In the positive recognition scenario, the goal is user authentication, wherein the user cooperates with the recognition system in order to be recognized and obtain access to locations or systems. In contrast, the negative recognition scenario deals with an uncooperative user who is unwilling to be identified.
Based on the recognition goal, the quality of the fingerprint can lead to different consequences.
In the positive recognition scenario, low-quality fingerprints prevent legitimate users from being authenticated. Although this brings inconvenience, users learn to reduce distortion after several authentication attempts. 
Serious consequences of low-quality fingerprints are tied with the negative recognition scenario in which users may deliberately decrease the quality of fingerprint to avoid being identified \cite{wein2005}. Actually, attempts of altering and damaging fingerprints in order to impair identification have been reported by law enforcement officials \cite{feng2010, yoon2012}. 
Hence,  increasing fingerprint quality is a necessary task in negative recognition systems. Additionally, it provides the added benefit of reducing the inconvenience of false rejection of valid users in positive recognition systems.

The quality of fingerprint samples can be deteriorated by many factors, either geometrically or photometrically. The primary cause of photometric degradation is artifacts on the finger or sensor, such as oil, moisture or markings from previous impressions. Photometric degradation in fingerprints has been widely investigated in terms of detection \cite{alonso2007,  fierrez2006, tabassi2009} and compensation \cite{chikkerur2007, feng2013, hong1998, turroni2012, yang2014}.

Fingers have cylindrical shape with relatively small radius compared to ridge pattern size. Capturing fingerprint samples is a complex mapping from a 3D surface to a 2D image, since the finger is being pressed onto a platen on a sensor. This mapping differs for each impression, referred to as geometric distortion. Geometric distortion is related to mechanical properties, such as the force and torque a user applies to the finger in the acquisition process. Different from photometric distortion, geometric distortion introduces translational and rotational error in the relative distances and orientations of local features. These relative distances and orientations of local features are the abstract identifiers of a user. In the presence of photometric distortion, the match score decreases since many minutiae may be missing, or false minutiae may be detected. On the contrary, in cases of severe geometric distortion, the match score decreases because the new composition of minutiae forms a completely different ID caused by the distortion. The issue is more critical in negative recognition systems, since distorted samples are still of high quality compared to clean samples, but matching algorithms fail to recognize them.

In this paper, we address the geometric distortion problem of fingerprint recognition systems by proposing a fast and effective distortion estimator which captures the non-linear properties of geometric distortion of fingerprints. While recently proposed methods handle distortion using a dictionary of distorted templates, for this work, we use a DCNN to estimate the principal distortion components of input samples. Our approach has the following contributions: 
\begin{itemize}
  \item There is no need to estimate the ridge frequency and orientation maps of input fingerprints. 
  \item Distortion parameters are being estimated continuously to achieve more accurate rectifications.
  \item A notable decrease in rectification time due to embedding distortion templates in network parameters.
\end{itemize}
	
The rest of the paper is organized as follows. In section \ref{relatedwork}, related works are reviewed. Section \ref{proposedmethod} describes the proposed approach, and section \ref{experiments} presents the experimental results. Finally, we conclude the paper in section \ref{conclusion}.

\begin{figure*}
\begin{center}
\includegraphics[scale=0.3]{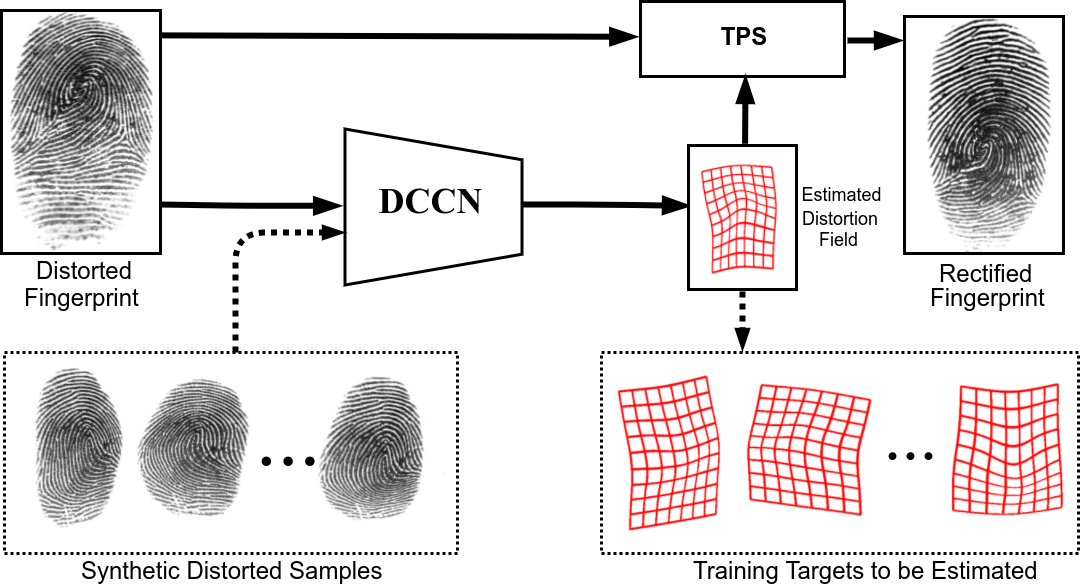}
\end{center}
   \caption{Flowchart of the proposed method for rectifying distorted fingerprints. The solid line shows testing path and the dash line shows training path.}
\label{fig:flowchart}
\end{figure*}
\section{Related Work}
\label{relatedwork}


Various approaches have been proposed in the literature to tackle the issue of geometric distortion in fingerprints. Designing specific acquisition hardware which detects distortion during recording procedure is a well-established approach. In this approach, the hardware detects distorted samples using different techniques, such as measuring excessive force \cite{bolle2000} or the  deformation of the acquisition surface \cite{fujii2010}, and motion processing during capturing fingerprint video \cite{dorai2004}. The hardware rejects severely distorted records and asks the user to provide a new impression until the system requirements are satisfied. Despite the improvements in recognition performance [16], there are certain drawbacks associated with the use of hardware-based distortion detection techniques: (i) they need specific sensors and additional capabilities; (ii) it is not possible to apply them on previously recorded samples; (iii) it makes the system weak against malicious users who have altered their finger tips and ridge patterns; (iv) it is merely detecting distortion, and there is no rectification process since user is obligated to provide clean impressions.
	
Since geometric distortion essentially moves features in fingerprints, adding distortion tolerance to fingerprint matching has shown promising results in compensating for the distortion problem \cite{ratha1996, chen2006, bazen2003, thebaud1999, kovacs2000, feng2006}. 
Distortion can be modeled by different special transformations such as rigid and thin plate spline (TPS) \cite{bookstein1989}. Although rigid transformation is not powerful enough to model the complex properties of geometric distortion, combining a global rigid transform and a local tolerant window have shown improvements in matching distorted samples \cite{ratha1996, chen2006}. TPS as a more complex transformation has been used to make matching algorithms tolerant to  geometric distortion \cite{bazen2003}.
However, compensating for distortion by adding tolerance to a fingerprint matcher inevitably results in a higher false positive match rate, and is highly dependent on estimating parameters of a complex transformation function. 

Ross et al. \cite{ross2005, ross2006} proposed a rectification technique based on learning deformation pattern from the correspondence of ridge curvatures of the same finger in different impressions. 
By computing average distortion based on corresponding ridges, it is possible to estimate parameters of the TPS transformation. This method showed improvement in matching distorted samples. However, the performance of the ridge curve correspondence method is highly dependent on the number of impressions of the same finger, and in most databases there are not enough samples per class to provide such an estimation.

Based on the assumption that the ridge frequency within a normal fingerprint is constant, Senior and Bolle \cite{senior2001} introduced a mathematical method of distortion rectification by equalizing the frequency map in distorted fingerprints.
Their method improves matching performance, especially when applying equalization to both distorted and original samples before matching.
Although it has been shown in \cite{wan2006, feng2008} that the ridge frequency map has discriminative information, and clearly it is not constant within the whole fingerprint area, their approach offered two important accomplishments compared to previous works. First, it does not need any specific hardware design, and second, it is possible to apply their algorithm on a single fingerprint image. However, equalizing all ridge spacings in a fingerprint has the following limitations: (i) some identification information will be lost and the false positive match rate will increase; (ii) in severe distortion cases, ridges are mixed together, and it is not possible to equalize the spacing between them; and (iii) equalizing the ridge frequency map within the whole fingerprint introduces distortion in the ridge orientation map.

More recently, Si et al. \cite{ref1} collected the Tsinghua distorted fingerprint database by inducing 10 different types of force and torque to fingers during the fingerprint acquisition process. They proposed a statistical model for distortion by computing minutiae displacements in distorted and corresponding original samples. In this method, the top two significant principal components of displacement are used to generate a dictionary of distorted samples. For each input sample, the ridge frequency and orientation maps are computed and compared to a dictionary in order to find the nearest distorted template. Their method shares all advantages of previous works, and it does not equalize the ridge frequency map. Therefore, discriminatory information of the frequency map is preserved and the ridge orientation map is not distorted. Considering all advantages of using a dictionary of distorted templates, there are still some limitations that need to be addressed: (i) computing frequency and orientation maps for input samples and comparing them with all samples in the dictionary takes a significant amount of time (from a second to several minutes depending on fingerprint properties); (ii) the performance of this method is related to the dictionary size, and increasing the dictionary size makes system slower; and iii) this method is highly dependent on computing the frequency and orientation maps of input samples which are not reliable due to the presence of distortion. 



\begin{table*}[]
\centering

\begin{tabular}{|c|c|l|l|l|l|}
\hline
\textbf{Layer} & \textbf{Type}  & \multicolumn{2}{c|}{\textbf{Kernel Size}} & \multicolumn{1}{c|}{\textbf{Input Size}} & \multicolumn{1}{c|}{\textbf{Output Size}} \\ \hline
1                         & Conv, BN, ReLU, MP        & \multicolumn{2}{l|}{$3\times3\times32$}     & $256 \times 256\times1$                    & $128\times128\times32$                     \\ \hline
2                         & Conv, BN, ReLU, MP        & \multicolumn{2}{l|}{$3\times3\times64$}     & $128 \times 128\times32$                   & $64\times64\times64$                     \\ \hline
3                         & Conv, BN, ReLU, MP        & \multicolumn{2}{l|}{$3\times3\times64$}     & $64\times64 \times64$                      & $32\times32\times64$                    \\ \hline
4                         & Conv, BN, ReLU, MP        & \multicolumn{2}{l|}{$3\times3\times128$}    & $32\times32\times64$                       & $16\times16\times 128$                     \\ \hline
5                         & Conv, BN, ReLU, MP        & \multicolumn{2}{l|}{$3\times3\times256$}    & $16\times16\times128$                      & $8\times8\times256$                       \\ \hline
6                         & Conv, BN, ReLU, MP        & \multicolumn{2}{l|}{$3\times3\times512$}    & $8\times8\times256$                         & $4\times4\times512$                          \\ \hline
7                         & Conv, BN, ReLU, MP        & \multicolumn{2}{l|}{$3\times3\times1024$}   & $4\times4\times512$                         & $2\times2\times1024$                        \\ \hline
8                         & Conv, BN, ReLU, MP        & \multicolumn{2}{l|}{$3\times3\times2048$}   & $2\times2\times1024$                         & $1\times1\times2048$                        \\ \hline
9                         & \multicolumn{1}{l|}{Conv} & \multicolumn{2}{l|}{$1\times1\times2$}      & $1\times1\times2048$                        & $1\times1\times2$                        \\ \hline
\end{tabular}
\vspace{5pt}
\caption{Architecture of the proposed DCNN used for estimating the distortion fields. All layers except the last one comprise Convolution (Conv), Batch Normalization (BN), ReLU and Max Pool (MP). All max poolings are $2\times2$ with the stride of two. All convolution strides are one, and all inputs to convolutions are padded to have the same size outputs.}
\label{tab:arch}
\end{table*}

\section{DCNN-based Distortion Estimation Model}
\label{proposedmethod}

Our method is inspired by the rectification approach proposed by Si et al. \cite{ref1, ref2}. The major limitation of their method is related to identifying the nearest distorted template in a dictionary of distorted samples. Finding the nearest neighbor to the distorted input sample in the dictionary is not accurate due to unreliable frequency and orientation maps extracted from the input sample. Instead of using a dictionary of the ridge frequency and orientation maps of distortion templates, we use a DCNN to estimate distortion parameter of the input sample. In this way, the non-linear transformations that caused distorted templates are being learned by the deep neural network during the training phase. The input to the network is the raw fingerprint image, and there is no need for computing the ridge frequency and orientation maps for the input samples. Contrary to the dictionary-based approach, the computational time of our proposed DCNN for estimating the distortion for an input, does not change by increasing the number of training samples since the network has a fixed number of parameters. On the other hand, the DCNN is capable of learning complex combinations of geometric distortions. A flowchart depicting the rectification scheme of the proposed method is shown in Figure \ref{fig:flowchart}. In the training phase, the network learns to estimate the distortion parameters of the input training images by minimizing the difference between the estimated parameters and the actual values. In the testing phase, the network estimates distortion parameters by mapping the input fingerprint to a non-linear manifold of distortion bases. Using the estimated distortion template and the input fingerprint, it is possible to rectify the distorted fingerprint by the inverse TPS \cite{bookstein1989} transformation of the distortion.

\subsection{Modeling Geometric Distortion to Generate Synthetic Distorted Fingerprints }

\begin{figure}
\begin{center}
\includegraphics[scale=.3]{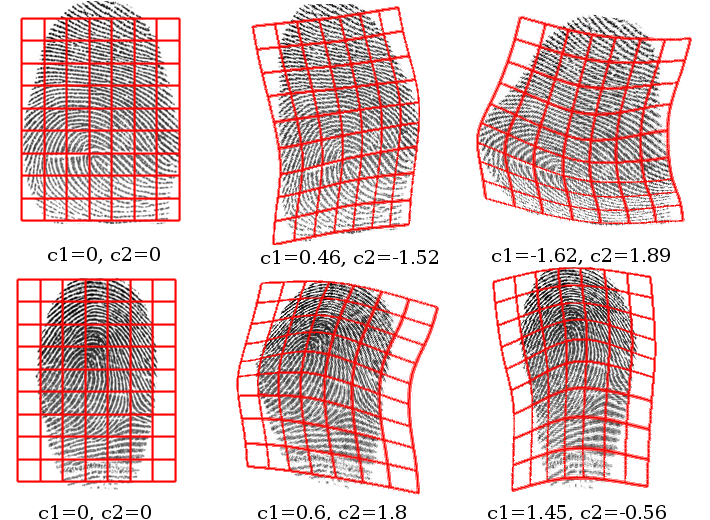}
\end{center}
   \caption{Examples of synthetic distorted fingerprint samples generated for training the network. Each sample is generated by randomly sampling distortion bases $c_1$, $c_2$.}
\label{fig:synsamples}
\end{figure}

Training a DCNN requires a comprehensive database of labeled images. We generated a synthetic database of distorted images in order to train our network. It is essential to model distortion for this purpose. Similar to \cite{ref1}, we used the Tsinghua distorted fingerprint database to statistically model geometric distortion. To extract displacement due to geometric distortion, we matched minutiae pairs from the original and distorted fingerprint samples. Minutia detection was performed using VeriFinger 7.0 SDK \cite{verifinger}. Since minutiae are anomalies in the fingerprint ridge map and have random positions we defined a similar grid of points as in \cite{ref1} to have a reference of distortion to be compared among different fingers. Using sampling grid pairs from the original and distorted fingerprints, it is possible to represent distortion as a displacement of corresponding points on the original grid and the distorted grid as follows:
\begin{equation}
d_i = x_i^D - x_i^N,
\end{equation}
where $d_i$ is the displacement of minutia for the $i$th pair of distorted and the corresponding normal fingerprint. Using distortion samples of the Tsinghua database and computing the distortion fields, it is possible to statistically model distortion by its principal components using PCA \cite{novikov2005, tang2009, rueckert2003}. Approximation of distortion fields using PCA will be:
\begin{equation}
\hat{d} 	\approx \overline{d} + \sum_{i=1}^{t}c_i \sqrt{\lambda_i}e_i.
\end{equation}
In the above equation, $t$ is the number of selected principal components, $c_i$ is the coefficient of the corresponding eigenvector component, $e_i$ is $i$th eigenvector and $\lambda_i$ is its corresponding eigenvalue. We used the first two significant eigenvectors of distortion to generate our synthetic samples. We generated a dataset of synthetic distorted fingerprints using 1033 normal fingerprints from the BioCOP 2013 dataset \cite{biocop}. Each normal fingerprint was transformed to 400 distorted images by sampling each of the two principal distortion components extracted from the Tsinghua database. Sampling was performed randomly with a uniform distribution between -2 and 2. The generated dataset has $1033 \times 401 = 414,233$ samples, in which each ID has one normal sample and 400 distorted samples. Figure \ref{fig:synsamples} shows two generated samples for two different fingers.


\begin{figure*}[t]
\begin{center}
\includegraphics[scale=.29]{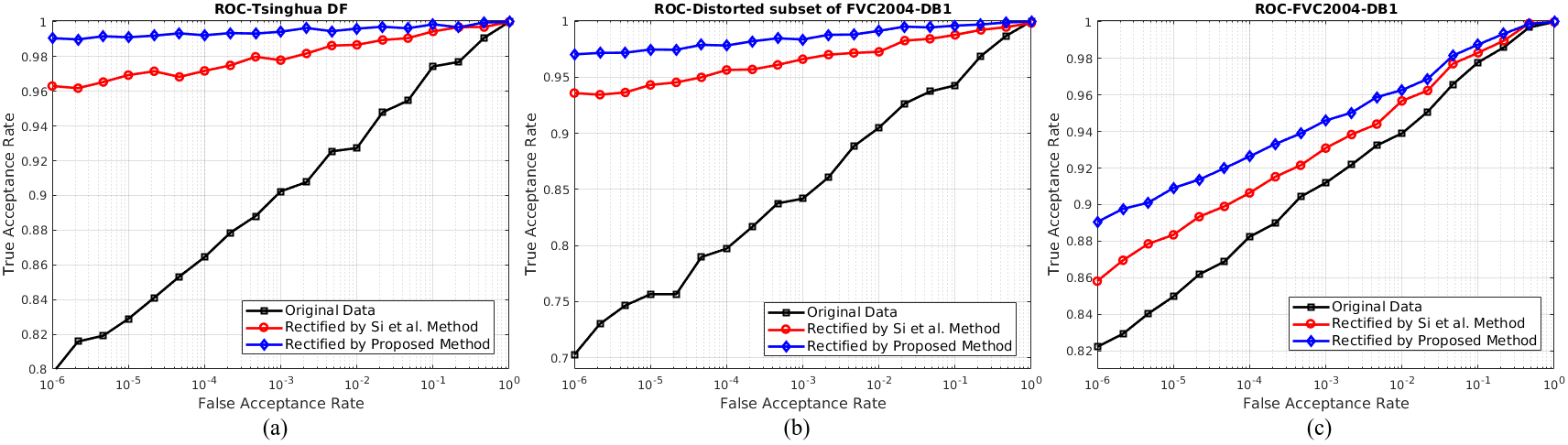}
\end{center}
   \caption{The ROC curves of three matching experiments for the following three databases (a) Tsinghua DF database, (b) FVC2004 DB1 and (c) geometrically distorted subset of FVC2004 DB1. }
\label{fig:roc}
\end{figure*}

\subsection{Network Architecture}

We used a deep convolutional neural network to learn the two eigenvector-based distortion coefficients. Compared to the fully connected networks, DCNNs are more robust against over-fitting due to weight sharing and fewer learning parameters. All layers except the last one are convolutional layers. The input image to the network has a size of $256\times256\times1$ pixels (first dimension is width, second is height and third is the depth). Our network consists of 9 convolutional blocks. Each layer, except the last one, comprises convolution, batch normalization, Rectified Linear Unit (ReLU) and max polling with stride equal to two. A detailed properties of the network is shown in Table \ref{tab:arch}. 

The network minimizes the norm-2 distance between ground truth coefficients ($c_1$ and $c_2$) and the DCNN outputs. For training the model, we first centered images according to the center of mass of the fingerprint area, and then scaled and cropped inputs to a size of $256\times256$. We used 401,000 synthetic distorted fingerprint images to train the model. The network was trained over 40 epochs, each epoch consisting of 6,265 iterations with a batch size = 64. Adam optimization method \cite{kingma2014, svoboda2017} is used as the optimizer due to its fast convergence with beta = 0.5 and learning rate = $10^{-4}$.


\section{Experiments}
\label{experiments}

\begin{table}[]
\centering
\begin{tabular}{c|c|c|}
\cline{2-3}
\multicolumn{1}{l|}{}           &       \multicolumn{2}{c|}{Time (sec)}           \\ \hline
\multicolumn{1}{|c|}{Method}    &       Tsinghua DF    & FVC2004 DB1              \\ \hline
\multicolumn{1}{|c|}{Si et al. \cite{ref1}}            & 8.373     & 7.816          \\ \hline
\multicolumn{1}{|c|}{Our}       & 0.741           & 0.736                         \\ \hline
\end{tabular}
\vspace{2mm}
\caption{Average time of distortion estimation. The proposed DCNN distortion estimation method is approximately 10 times faster than the nearest neighbor method used by Si et al. \cite{ref1}.}
\label{asdf}
\end{table}

\begin{figure*}
\begin{center}
\includegraphics[scale=.45]{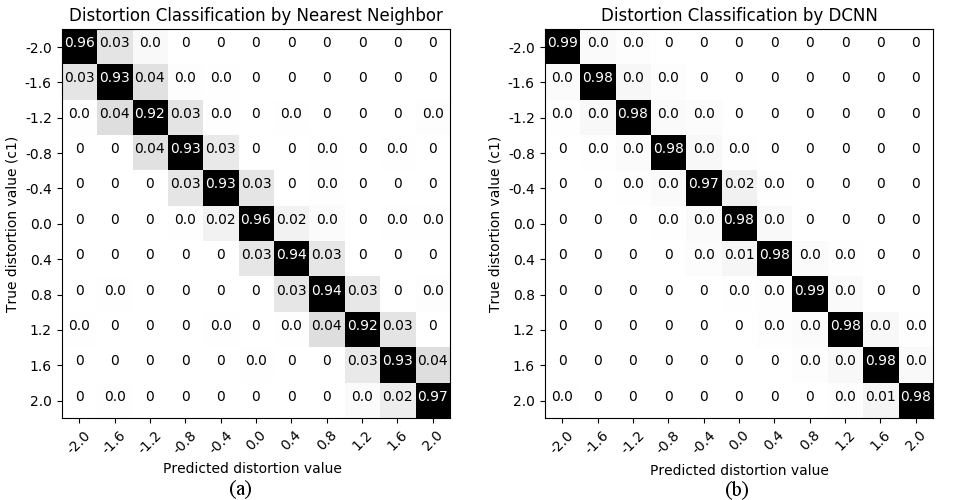}
\end{center}
   \caption{Confusion matrices for the following approaches (a) the nearest neighbor method by Si et al. \cite{ref1} and (b) the proposed DCNN-based distortion estimation.}
\label{fig:cm}
\end{figure*}

\begin{figure*}
\begin{center}
\includegraphics[scale=0.35]{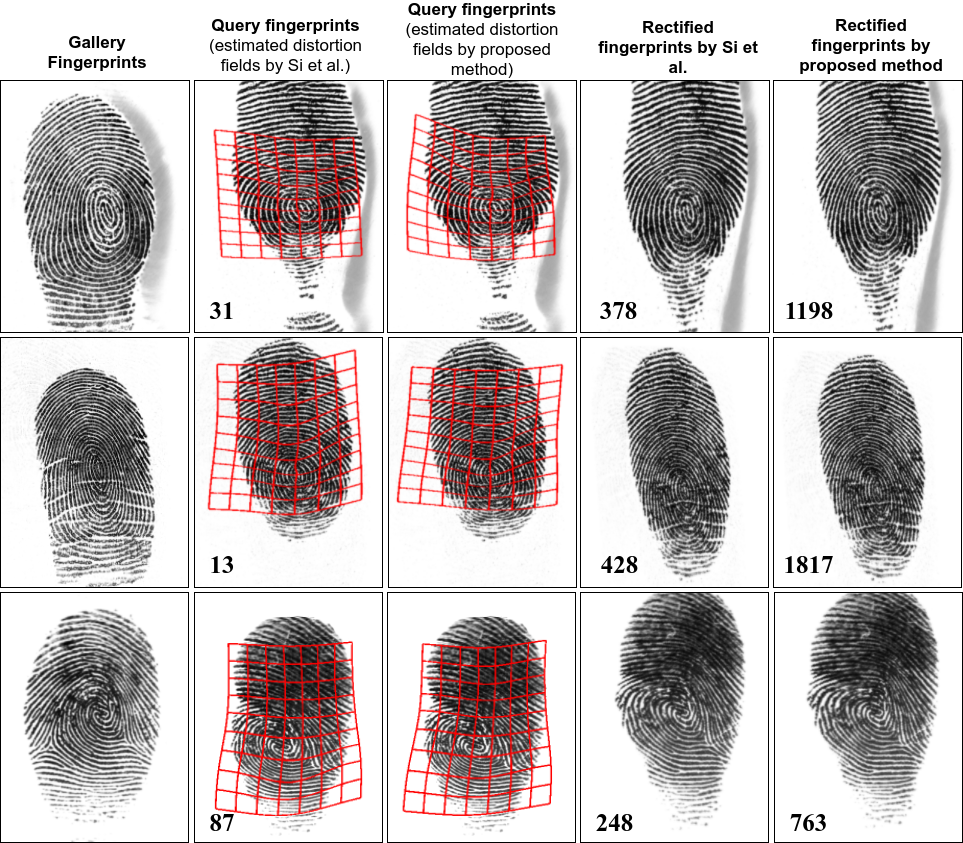}
\end{center}
   \caption{Match scores for three pairs of normal and rectified fingerprints by two different approaches. The red grid on query fingerprints shows estimated distortion fields by our method and the method proposed by Si et al. \cite{ref1}. Two first samples are from the Tsinghua DF database and the third sample is from FVC2004 DB1.}
\label{fig:compare}
\end{figure*}

Our first performance measure for evaluating the proposed distortion rectification is the overall matching performance. To evaluate the contribution of the proposed method in improving matching performance, we conducted three experiments on each of the following three databases: FVC2004 DB1, distorted subset of FVC2004 DB1 and Tsinghua DF database. VeriFinger 7.0 SDK \cite{verifinger} is used to match fingerprint samples. 

The match score in each experiment is calculated for pairs of samples with the same ID, and no imposter pairs are conducted since the match score of VeriFinger is linked to the false acceptance rate (FAR). Higher match scores have a lower chance of falsely being accepted.
In all three matching experiments, the first sample in each pair is a normal fingerprint without distortion, and the second one is the original distorted sample or the rectified sample. Rectification is performed both by our method and the method proposed by Si et al. \cite{ref1}. ROC curves on three databases are depicted in Figure \ref{fig:roc}.

In the first experiment, samples from the Tsinghua DF database are rectified to evaluate the training procedure of the network and the rectification performance. The Tsinghua DF database consists of 320 pairs of normal and distorted fingerprints from 185 different fingers. 

Network training is performed using a synthetic distorted dataset generated by randomly sampling the first two significant principal components of the distortion manifold extracted from the Tsinghua DF database. Although the network has never seen the original samples from the Tsinghua DF database during the training procedure, distortion components used to generate the synthetic dataset may bias the performance of the network. 
Therefore, it is essential to evaluate matching performance on a dataset containing only geometric distortion that is different from the Tsinghua DF database. In the second experiment, a geometrically distorted subset of FVC2004 DB1 is used to evaluate the rectification performance of the proposed method. The distorted subset of FVC2004 DB1 contains 89 samples with skin distortions.   

In the third experiment, FVC2004 DB1 is used to evaluate the rectification performance on a distorted database containing a variety of geometric and photometric distortions. FVC2004 DB1 consists of 110 classes and eight samples per class. Samples of each class are acquired by deliberately inducing photometric or geometric distortions. Since FVC2004 DB1 contains different distortion types, the proposed method targets only geometrically distorted samples and rejects other distortion types.

The quality of rectified distorted samples depends on the performance of the distortion estimation algorithm. We conducted an experiment to compare distortion estimation of DCNN with the nearest neighbor method used by Si et al. \cite{ref1}. The synthetic distorted database used in this paper was generated using random sampling of the first two significant principal components. For comparison purposes, we generated another distorted database that was the same as Si et al. \cite{ref1} to compare distortion classification of the two methods. The proposed DCNN estimates continuous values of distortion basis. Therefore, we quantized the network output to have 11 classes for each basis. In this order, class 1 is the first distortion basis with coefficient equal to -2.0, and class 11 is the first distortion basis with coefficient equal to 2.0. The confusion matrices for the two methods of classifying the first basis are shown in Figure \ref{fig:cm}. The Distribution of diagonal values of the second confusion matrix shows that the proposed DCNN is much more precise in estimating distortion coefficients. Although nearest neighbor is not accurate enough, it contributes to distortion rectification since it finds the target distortion class with an error margin of approximately two classes. 

To compare the rectification results of our approach and the method proposed by Si et al. \cite{ref1}, three examples from the Tsinghua DF database and FVC2004 DB1 are shown in Figure \ref{fig:compare}. The rectified samples by both methods are very similar but the match score measurement indicates that there is a significant difference between them. A slight estimation error in distortion parameters prevents the spatial transformation from correctly restoring minutiae displacements.

In a fingerprint recognition system, distortion rectification is one of the preprocessing steps that can affect the total response time of the system. It is not possible nor efficient to use a computationally slow rectification method in a real-time recognition system since it brings inconvenience to users. Therefore, it is essential to evaluate the rectification speed. We conducted two experiments to evaluate the average response time of the rectification process on a PC with 3.3 GHz CPU and NIVDIA TITAN X GPU. Results are reported in Table \ref{asdf}. From the average response time of the proposed approach and the matching experiments, it can be observed that the proposed DCNN as a distortion estimator, not only increases the accuracy of distortion detection, but also significantly reduces the detection time. 

An important fact to be considered is that the proposed algorithm is executed on the GPU, but the nearest neighbor method is executed on the CPU because it is not possible to implement a search method on parallel processors. Therefore, the reduction of the rectification time is mainly because of the capability of neural networks to embed training samples in the network parameters which enables us to convert a search problem to a direct prediction problem.  

Additionally, contrary to the nearest neighbor method, the response time of the proposed DCNN is independent of the properties of input samples to the network, and guarantees an efficient lower bound for processing speed.  

\section{Conclusion}
\label{conclusion}
Geometric distortion significantly reduces the match score produced by a fingerprint verification system. In the positive recognition scenario, this causes inconvenience for users, but in the negative recognition scenario where users may intentionally distort their fingerprint, this can be considered as a security vulnerability. Therefore, it is essential to implement distortion rectification in order to prevent malicious users from hiding their identity, as well as reduce the inconvenience of using identification systems in authentication tasks. We proposed a novel approach to estimate distortion parameters from raw fingerprint images without computing the ridge frequency and orientation maps. A deep convolutional neural network is  utilized to estimate distortion parameters of input samples. We successfully rectified distorted samples from the Tsinghua DF database and FVC2004 DB1 using the estimated distortion template. A comprehensive database of distorted samples was generated in order to train our deep neural network. The experimental results on several databases showed that the DCNN can estimate the non-linear distortions of samples more accurately. Comparing to the previous works, our method decreased rectification time significantly by embedding the training samples in the network parameters. In addition, since the estimation time of the proposed method is independent of the training size, it is possible to increase the number of principal components which are used to generate the synthetic distorted database for the future works.

\begin{center}
ACKNOWLEDGEMENT
\end{center}

This work is based upon a work supported by the Center for Identification Technology Research and the National Science Foundation under Grant $\#1650474$

{\small
\bibliographystyle{ieee}
\bibliography{egbib}

\begin{thebibliography}{10}\itemsep=-1pt

\bibitem{alonso2007}
F.~Alonso-Fernandez, J.~Fierrez, J.~Ortega-Garcia, J.~Gonzalez-Rodriguez,
  H.~Fronthaler, K.~Kollreider, and J.~Bigun.
\newblock A comparative study of fingerprint image-quality estimation methods.
\newblock {\em IEEE Transactions on Information Forensics and Security},
  2(4):734--743, 2007.

\bibitem{bazen2003}
A.~M. Bazen and S.~H. Gerez.
\newblock Fingerprint matching by thin-plate spline modelling of elastic
  deformations.
\newblock {\em Pattern Recognition}, 36(8):1859--1867, 2003.

\bibitem{bolle2000}
R.~M. Bolle, R.~S. Germain, R.~L. Garwin, J.~L. Levine, S.~U. Pankanti, N.~K.
  Ratha, and M.~A. Schappert.
\newblock System and method for distortion control in live-scan inkless
  fingerprint images, May~16 2000.
\newblock US Patent 6,064,753.

\bibitem{bookstein1989}
F.~L. Bookstein.
\newblock Principal warps: Thin-plate splines and the decomposition of
  deformations.
\newblock {\em IEEE Transactions on pattern analysis and machine intelligence},
  11(6):567--585, 1989.

\bibitem{FVC2006}
R.~Cappelli, M.~Ferrara, A.~Franco, and D.~Maltoni.
\newblock Fingerprint verification competition 2006.
\newblock {\em Biometric Technology Today}, 15(7):7--9, 2007.

\bibitem{cappelli2006}
R.~Cappelli, D.~Maio, D.~Maltoni, J.~L. Wayman, and A.~K. Jain.
\newblock Performance evaluation of fingerprint verification systems.
\newblock {\em IEEE transactions on pattern analysis and machine intelligence},
  28(1):3--18, 2006.

\bibitem{chen2006}
X.~Chen, J.~Tian, and X.~Yang.
\newblock A new algorithm for distorted fingerprints matching based on
  normalized fuzzy similarity measure.
\newblock {\em IEEE Transactions on Image Processing}, 15(3):767--776, 2006.

\bibitem{chikkerur2007}
S.~Chikkerur, A.~N. Cartwright, and V.~Govindaraju.
\newblock {Fingerprint enhancement using STFT analysis}.
\newblock {\em Pattern recognition}, 40(1):198--211, 2007.

\bibitem{dorai2004}
C.~Dorai, N.~K. Ratha, and R.~M. Bolle.
\newblock Dynamic behavior analysis in compressed fingerprint videos.
\newblock {\em IEEE transactions on circuits and systems for video technology},
  14(1):58--73, 2004.

\bibitem{feng2008}
J.~Feng.
\newblock Combining minutiae descriptors for fingerprint matching.
\newblock {\em Pattern Recognition}, 41(1):342--352, 2008.

\bibitem{feng2010}
J.~Feng, A.~K. Jain, and A.~Ross.
\newblock Detecting altered fingerprints.
\newblock In {\em Pattern Recognition (ICPR), 2010 20th International
  Conference on}, pages 1622--1625. IEEE, 2010.

\bibitem{feng2006}
J.~Feng, Z.~Ouyang, and A.~Cai.
\newblock Fingerprint matching using ridges.
\newblock {\em Pattern Recognition}, 39(11):2131--2140, 2006.

\bibitem{feng2013}
J.~Feng, J.~Zhou, and A.~K. Jain.
\newblock Orientation field estimation for latent fingerprint enhancement.
\newblock {\em IEEE transactions on pattern analysis and machine intelligence},
  35(4):925--940, 2013.

\bibitem{fierrez2006}
J.~Fierrez-Aguilar, Y.~Chen, J.~Ortega-Garcia, and A.~K. Jain.
\newblock Incorporating image quality in multi-algorithm fingerprint
  verification.
\newblock In {\em ICB}, pages 213--220. Springer, 2006.

\bibitem{fujii2010}
Y.~Fujii.
\newblock Detection of fingerprint distortion by deformation of elastic film or
  displacement of transparent board, Feb.~9 2010.
\newblock US Patent 7,660,447.

\bibitem{hong1998}
L.~Hong, Y.~Wan, and A.~Jain.
\newblock Fingerprint image enhancement: Algorithm and performance evaluation.
\newblock {\em IEEE transactions on pattern analysis and machine intelligence},
  20(8):777--789, 1998.

\bibitem{kingma2014}
D.~Kingma and J.~Ba.
\newblock Adam: A method for stochastic optimization.
\newblock {\em arXiv preprint arXiv:1412.6980}, 2014.

\bibitem{kovacs2000}
Z.~M. Kovacs-Vajna.
\newblock A fingerprint verification system based on triangular matching and
  dynamic time warping.
\newblock {\em IEEE Transactions on Pattern Analysis and Machine Intelligence},
  22(11):1266--1276, 2000.

\bibitem{verifinger}
{Neurotechnology Inc., Verifinger}.
\newblock http://www.neurotechnology.com.

\bibitem{novikov2005}
S.~Novikov and O.~Ushmaev.
\newblock Principal deformations of fingerprints.
\newblock In {\em Audio-and Video-Based Biometric Person Authentication}, pages
  229--237. Springer, 2005.

\bibitem{ratha1996}
N.~K. Ratha, K.~Karu, S.~Chen, and A.~K. Jain.
\newblock A real-time matching system for large fingerprint databases.
\newblock {\em IEEE Transactions on Pattern Analysis and Machine Intelligence},
  18(8):799--813, 1996.

\bibitem{ross2005}
A.~Ross, S.~Dass, and A.~Jain.
\newblock A deformable model for fingerprint matching.
\newblock {\em Pattern Recognition}, 38(1):95--103, 2005.

\bibitem{ross2006}
A.~Ross, S.~C. Dass, and A.~K. Jain.
\newblock Fingerprint warping using ridge curve correspondences.
\newblock {\em IEEE Transactions on Pattern Analysis and Machine Intelligence},
  28(1):19--30, 2006.

\bibitem{rueckert2003}
D.~Rueckert, A.~F. Frangi, and J.~A. Schnabel.
\newblock {Automatic construction of 3-D statistical deformation models of the
  brain using nonrigid registration}.
\newblock {\em IEEE transactions on medical imaging}, 22(8):1014--1025, 2003.

\bibitem{senior2001}
A.~W. Senior and R.~M. Bolle.
\newblock Improved fingerprint matching by distortion removal.
\newblock {\em IEICE Transactions on Information and Systems}, 84(7):825--832,
  2001.

\bibitem{ref2}
X.~Si, J.~Feng, and J.~Zhou.
\newblock Detecting fingerprint distortion from a single image.
\newblock In {\em 2012 IEEE International Workshop on Information Forensics and
  Security (WIFS)}, pages 1--6, Dec 2012.

\bibitem{ref1}
X.~Si, J.~Feng, J.~Zhou, and Y.~Luo.
\newblock Detection and rectification of distorted fingerprints.
\newblock {\em IEEE Transactions on Pattern Analysis and Machine Intelligence},
  37(3):555--568, March 2015.

\bibitem{svoboda2017}
J.~Svoboda, F.~Monti, and M.~M. Bronstein.
\newblock Generative convolutional networks for latent fingerprint
  reconstruction.
\newblock {\em arXiv preprint arXiv:1705.01707}, 2017.

\bibitem{tabassi2009}
E.~Tabassi and P.~Grother.
\newblock Fingerprint image quality.
\newblock In {\em Encyclopedia of Biometrics}, pages 482--490. Springer, 2009.

\bibitem{tang2009}
S.~Tang, Y.~Fan, G.~Wu, M.~Kim, and D.~Shen.
\newblock Rabbit: rapid alignment of brains by building intermediate templates.
\newblock {\em NeuroImage}, 47(4):1277--1287, 2009.

\bibitem{thebaud1999}
L.~R. Thebaud.
\newblock Systems and methods with identity verification by comparison and
  interpretation of skin patterns such as fingerprints, June~1 1999.
\newblock US Patent 5,909,501.

\bibitem{turroni2012}
F.~Turroni, R.~Cappelli, and D.~Maltoni.
\newblock Fingerprint enhancement using contextual iterative filtering.
\newblock In {\em Biometrics (ICB), 2012 5th IAPR International Conference on},
  pages 152--157. IEEE, 2012.

\bibitem{wan2006}
D.~Wan and J.~Zhou.
\newblock Fingerprint recognition using model-based density map.
\newblock {\em IEEE Transactions on Image Processing}, 15(6):1690--1696, 2006.

\bibitem{wein2005}
L.~M. Wein and M.~Baveja.
\newblock Using fingerprint image quality to improve the identification
  performance of the us visitor and immigrant status indicator technology
  program.
\newblock {\em Proceedings of the National Academy of Sciences of the United
  States of America}, 102(21):7772--7775, 2005.

\bibitem{biocop}
{WVU multimodal dataset, Biometrics and Identification Innovation Center}.
\newblock http://biic.wvu.edu/.

\bibitem{yang2014}
X.~Yang, J.~Feng, and J.~Zhou.
\newblock Localized dictionaries based orientation field estimation for latent
  fingerprints.
\newblock {\em IEEE transactions on pattern analysis and machine intelligence},
  36(5):955--969, 2014.

\bibitem{yoon2012}
S.~Yoon, J.~Feng, and A.~K. Jain.
\newblock Altered fingerprints: Analysis and detection.
\newblock {\em IEEE transactions on pattern analysis and machine intelligence},
  34(3):451--464, 2012.

\end{thebibliography}
}

\end{document}